\documentclass[11pt,a4paper]{article}
\usepackage[hyperref]{emnlp2018}
\usepackage{times}
\usepackage{latexsym}
\usepackage{amsmath}
\usepackage{multirow}
\usepackage{float}
\usepackage{booktabs}
\usepackage{subcaption}
\usepackage{graphicx}

\newcommand{\topline}{\toprule}
\newcommand{\midline}{\midrule}
\newcommand{\bottomline}{\bottomrule}
\usepackage{url}
\makeatletter

\usepackage{tabularx}
\usepackage{colortbl}
\usepackage{hhline}
\usepackage{arydshln}

\aclfinalcopy

\title{Temporal Information Extraction by Predicting Relative Time-lines}

\author{Artuur Leeuwenberg \and Marie-Francine Moens \\ Department of Computer Science\\
         KU Leuven, Belgium\\
          { \tt \{tuur.leeuwenberg, sien.moens\}@cs.kuleuven.be}
}

\date{}

\begin{document}
\maketitle

\begin{abstract}
The current leading paradigm for temporal information extraction from text consists of three phases: (1) recognition of events and temporal expressions, (2) recognition of temporal relations among them, and (3) time-line construction from the temporal relations. In contrast to the first two phases, the last phase, time-line construction, received little attention and is the focus of this work. In this paper, we propose a new method to construct a linear time-line from a set of (extracted) temporal relations. But more importantly, we propose a novel paradigm in which we directly predict start and end-points for events from the text, constituting a time-line without going through the intermediate step of prediction of temporal relations as in earlier work. Within this paradigm, we propose two models that predict in linear complexity, and a new training loss using TimeML-style annotations, yielding promising results.
\end{abstract}
\section{Introduction}

\begin{figure*}
\centering
\includegraphics[width=.9\textwidth]{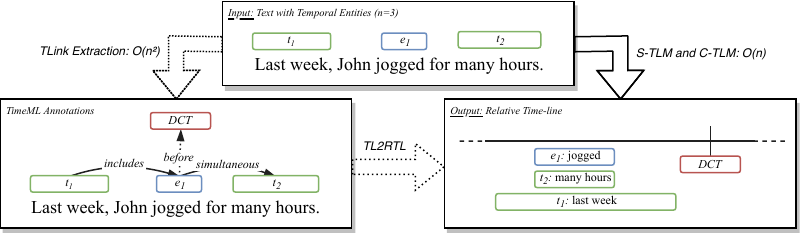}
\caption{\label{fig:example} An overview of two paradigms: (1) The indirect approach (dashed arrows), where first TLinks are predicted from which we can build a relative time-line using TL2RTL. And (2), the direct approach (solid arrow), where a relative time-line is predicted directly from the input by S-TLM or C-TLM. } 
\end{figure*}

The current leading perspective on temporal information extraction regards three phases: (1) a \textit{temporal entity recognition} phase, extracting events (blue boxes in Fig. \ref{fig:example}) and their attributes, and extracting temporal expressions (green boxes), and normalizing their values to dates or durations, (2) a \textit{relation extraction} phase, where temporal links (TLinks) among those entities, and between events and the document-creation time (DCT) are found (arrows in Fig. \ref{fig:example}, left). And (3), construction of a time-line (\mbox{Fig. \ref{fig:example}}, right) from the extracted temporal links, if they are temporally consistent. Much research concentrated on the first two steps, but very little research looks into step 3, time-line construction, which is the focus of this work.

In this paper, we propose a new time-line construction paradigm that evades phase 2, the relation extraction phase, because in the classical paradigm temporal relation extraction comes with many difficulties in training and prediction that arise from the fact that for a text with $n$ temporal entities (events or temporal expressions) there are $n^2$ possible entity pairs, which makes it likely for annotators to miss relations, and makes inference slow as $n^2$ pairs need to be considered. Temporal relation extraction models consistently give lower performance than those in the entity recognition phase \cite{UzZaman2013,bethard2016semeval,Bethard2017SemEval-2017TempEval}, introducing errors in the time-line construction pipe-line.

The ultimate goal of our proposed paradigm is to predict from a text in which entities are already detected, for each entity: (1) a probability distribution on the entity's starting point, and (2) another distribution on the entity's duration. The probabilistic aspect is crucial for time-line based decision making. Constructed time-lines allow for further quantitative reasoning with the temporal information, if this would be needed for certain applications.




As a first approach towards this goal, in this paper, we propose several initial time-line models in this paradigm, that directly predict - in a linear fashion - start points and durations for each entity, using text with annotated temporal entities as input (shown in \mbox{Fig. \ref{fig:example}}). The predicted start points and durations constitute a \textit{relative time-line}, i.e. a total order on entity start and end points. The time-line is relative, as start and duration values cannot (yet) be mapped to absolute calender dates or durations expressed in seconds. It represents the relative temporal order and inclusions that temporal entities have with respect to each other by the quantitative start and end values of the entities. Relative time-lines are a first step toward our goal, building models that predict statistical absolute time-lines. To train our relative time-line models, we define novel loss functions that exploit TimeML-style annotations, used in most existing temporal corpora.

This work leads to the following contributions:
\begin{itemize}
\item A new method to construct a relative time-line from a set of temporal relations (TL2RTL).
\item Two new models that, for the first time, directly predict (relative) time-lines - in linear complexity - from entity-annotated texts without doing a form of temporal relation extraction (S-TLM \& C-TLM).
\item Three new loss functions based on the mapping between Allen's interval algebra and the end-point algebra to train time-line models from TimeML-style annotations.
\end{itemize}

In the next sections we will further discuss the related work on temporal information extraction. We will describe the models and training losses in detail, and report on conducted experiments.

\section{Related Work}
\label{sec:related_work}

\subsection{Temporal Information Extraction}
The way temporal information is conveyed in language has been studied for a long time. It can be conveyed directly through verb tense, explicit temporal discourse markers (e.g. \textit{during} or \textit{afterwards}) \cite{Derczynski2017} or temporal expressions such as dates, times or duration expressions (e.g. \textit{10-05-2010} or \textit{yesterday}). Temporal information is also captured in text implicitly, through background knowledge about, for example, duration of events mentioned in the text (e.g. even without context, \textit{walks} are usually shorter than \textit{journeys}).

Most temporal corpora are annotated with TimeML-style annotations, of which an example is shown in Fig \ref{fig:example}, indicating temporal entities, their attributes, and the TLinks among them.

The automatic extraction of TimeML-style temporal information from text using machine learning was first explored by \citet{Mani2006MachineRelations}. They proposed a multinomial logistic regression classifier to predict the TLinks between entities. They also noted the problem of missed TLinks by annotators, and experimented with using temporal reasoning (temporal closure) to expand their training data.

Since then, much research focused on further improving the pairwise classification models, by exploring different types of classifiers and features, such as (among others) logistic regression and support vector machines \cite{bethard2013cleartk,lin2015multilayered}, and different types of neural network models, such as long short-term memory networks (LSTM) \cite{Tourille2017,cheng2017classifying}, and convolutional neural networks (CNN) \cite{dligach2017neural}. Moreover, different sieve-based approaches were proposed \cite{chambers2014dense,Mirza2016a}, facilitating mixing of rule-based and machine learning components.

Two major issues shared by these existing approaches are: (1) models classify TLinks in a pairwise fashion, often resulting in an inference complexity of $O(n^2)$, and (2) the pair-wise predictions are made independently, possibly resulting in prediction of temporally inconsistent graphs. To address the second, additional temporal reasoning can be used at the cost of computation time, during inference \cite{chambers2008jointly,Denis2011,Do2012JointConstruction}, or during both training and inference \cite{yoshikawa2009jointly,Laokulrat2015,Ning2017,Leeuwenberg2017a}. In this work, we circumvent these issues, as we predict time-lines - in linear time complexity - that are temporally consistent by definition.

\subsection{Temporal Reasoning}

Temporal reasoning plays a central role in temporal information extraction, and there are roughly two approaches: (1) Reasoning directly with Allen's interval relations (shown in Table \ref{tab:relations}), by constructing rules like: If event X occurs before Y, and event Y before Z then X should happen before Z \cite{Allen}. Or (2), by first mapping the temporal interval expressions to expressions about interval end-points (start and endings of entities) \cite{vilain1990constraint}. An example of such mapping is that If event X occurs before Y then the end of X should be before the start of Y. Then reasoning can be done with end-points in a point algebra, which has only three point-wise relations ($=, <, >$), making reasoning much more efficient compared to reasoning with Allen's thirteen interval relations. 

Mapping interval relations to point-wise expressions has been exploited for model inference by \citet{Denis2011}, and for evaluation by \citet{UzZaman:2011}. In this work, we exploit it for the first time for model training, in our loss functions.





\section{Models}
\begin{figure*}
\centering
\includegraphics[width=\textwidth]{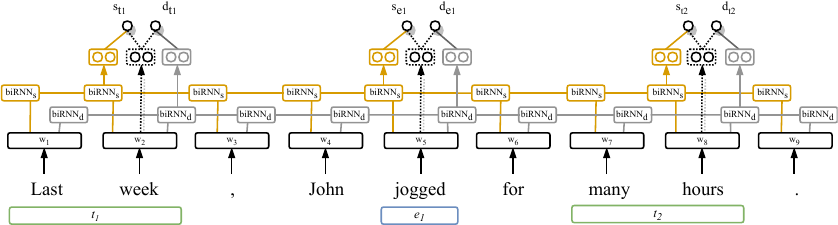}
\caption{\label{fig:pred_model} Schematic overview of our two time-line models: C-TLM (solid edges), exploiting entity context, and the simpler S-TLM (dotted edges), which is context independent. The models predict a starting point (s) and duration (d) for each given temporal entity ($t_1$, $e_1$, and $t_2$) in the input.}
\end{figure*}

We propose two model structures for direct time-line construction: (1) a simple context-independent model (S-TLM), and (2) a contextual model (C-TLM). Their structures are shown in Fig. \ref{fig:pred_model}. Additionally, we propose a method to construct relative time-lines from a set of (extracted) TLinks (TL2RTL).
In this section we first explain the first two direct models S-TLM and C-TLM, and afterwards the indirect method TL2RTL.

\subsection{Direct Time-line Models}
\subsubsection*{Word representation}

In both S-TLM and C-TLM, words are represented as a concatenation of a word embedding, a POS embedding, and a Boolean feature vector containing entity attributes such as the type, class, aspect, following \cite{Do2012JointConstruction}. Further details on these are given in the experiments section.


\subsubsection*{Simple Time-line Model (S-TLM)}
For the simple context-independent time-line model, each entity is encoded by the word representation of the last word of the entity (generally the most important). From this representation we have a linear projection to the duration $d$, and the start $s$. S-TLM is shown by the dotted edges in Fig \ref{fig:pred_model}. An advantage of S-TLM is that it has very few parameters, and each entity can be placed on the time-line independently of the others, allowing parallelism during prediction. The downside is that S-TLM is limited in its use of contextual information.

\subsubsection*{Contextual Time-line Model (C-TLM)}
To better exploit the entity context we also propose a contextual time-line model C-TLM (solid edges in Fig \ref{fig:pred_model}), that first encodes the full text using two bi-directional recurrent neural networks, one for entity starts (BiRNN$_s$), and one for entity durations (BiRNN$_d$).\footnote{We also experimented with sharing weights among BiRNN$_d$ and BiRNN$_s$. In our experiments, this gave worse performance, so we propose to keep them separate.}
On top of the encoded text we learn two linear mappings, one from the BiRNN$_d$ output of the last word of the entity mention to its duration $d$, and similarly for the start time, from the BiRNN$_s$ output to the entity's start $s$.

\subsubsection*{Predicting Start, Duration, and End}
Both proposed models use linear mappings\footnote{Adding more layers did not improve results.} to predict the start value $s_i$ and duration $d_i$ for the encoded entity $i$. By summing start $s_i$ and duration $d_i$ we can calculate the entity's end-point $e_i$. 
\begin{align}
\label{eq:end}
e_i = s_i + \max(d_i, d_{min})
\end{align}
Predicting durations rather than end-points makes it easy to control that the end-point lies after the start-point by constraining the duration $d_i$ by a constant minimum duration value $d_{min}$ above 0, as shown in Eq. \ref{eq:end}.

\subsubsection*{Modeling Document-Creation Time}
Although the DCT is often not found explicitly in the text, it is an entity in TimeML, and has TLinks to other entities. We model it by assigning it a text-independent start $s_{\textsc{dct}}$ and duration $d_{\textsc{dct}}$. 

Start  $s_{\textsc{dct}}$ is set as a constant (with value 0). This way the model always has the same reference point, and can learn to position the entities w.r.t. the DCT on the time-line. 

In contrast, DCT duration  $d_{\textsc{dct}}$ is modeled as a single variable that is learned (initialized with 1). Since multiple entities may be included in the DCT, and entities have a minimum duration $d_{min}$, a constant $d_{\textsc{dct}}$ could possibly prevent the model from fitting all entities in the DCT. Modeling $d_{\textsc{dct}}$ as a variable allows growth of $d_{\textsc{dct}}$ and averts this issue.\footnote{Other combinations of modeling $s_{\textsc{dct}}$ and $d_{\textsc{dct}}$ as variable or constant decreased performance.}

\subsubsection*{Training Losses}


We propose three loss functions to train time-line models from TimeML-style annotations: a regular time-line loss $L_\tau$, and two slightly expanded discriminative time-line losses, $L_{\tau ce}$ and $L_{\tau h}$. 

\subsubsection*{Regular Time-line Loss ($L_\tau$)}
Ground-truth TLinks can be seen as constraints on correct positions of entities on a time-line. The regular time-line loss $L_\tau$ expresses the degree to which these constraints are met for a predicted time-line. If all TLinks are satisfied in the time-line for a certain text, $L_\tau$ will be 0 for that text.

As TLinks relate entities (intervals), we first convert the TLinks to expressions that relate the start and end points of entities.
How each TLink is translated to its corresponding point-algebraic constraints is given in Table \ref{tab:relations}, following \citet{Allen}.
\begin{table}[h!]
\centering
\caption{Point algebraic interpretation ($I_{PA}$) of temporal links used to construct the loss function. The start and end points of event $X$ are indicated by $s_x$ and $e_x$ respectively.}
\label{tab:relations}
\resizebox{.49\textwidth}{!}{
\begin{tabular}{llc}\topline
\textbf{Allen Algebra} & \multicolumn{1}{l}{\textbf{Temporal Links}} & \textbf{Point Algebra} \\\midline
\begin{tabular}[c]{@{}l@{}}X precedes Y\\ Y preceded by X\end{tabular} & \begin{tabular}[c]{@{}l@{}}X before Y\\ Y after X\end{tabular} & $e_x < s_y$   \\\midline
\begin{tabular}[c]{@{}l@{}}X starts Y\\ Y started by X\end{tabular} & \begin{tabular}[c]{@{}l@{}}X begins Y\\ Y begun by X\end{tabular} & \begin{tabular}[c]{@{}l@{}}$s_x=s_y$\\ $e_x<e_y$\end{tabular}  \\\midline
\begin{tabular}[c]{@{}l@{}}X finishes Y\\ Y finished by X\end{tabular} & \begin{tabular}[c]{@{}l@{}}X ends Y\\ Y ended by X\end{tabular} & \begin{tabular}[c]{@{}l@{}}$e_x=e_y$\\ $s_y<s_x$\end{tabular}   \\\midline
\begin{tabular}[c]{@{}l@{}}X during Y\\ Y includes X\end{tabular} & \begin{tabular}[c]{@{}l@{}}X is included Y\\ Y includes X\end{tabular} & \begin{tabular}[c]{@{}l@{}}$s_y<s_x$\\ $e_x<e_y$\end{tabular}  \\\midline
\begin{tabular}[c]{@{}l@{}}X meets Y\\ Y met by X\end{tabular} & \begin{tabular}[l]{@{}l@{}}X immediately before Y\\ Y immediately after X\end{tabular} & $e_x=s_y$   \\\midline
\begin{tabular}[c]{@{}l@{}}X overlaps Y\\ Y overlapped by X\end{tabular} & \begin{tabular}[c]{@{}l@{}}absent\footnotemark[4]\\ absent\footnotemark[4]\end{tabular} & \begin{tabular}[c]{@{}l@{}}$s_x<s_y$\\ $s_y<e_x$\\ $e_x<e_y$\end{tabular}   \\\midline
X equals Y & \begin{tabular}[c]{@{}l@{}}X simultaneous Y\\ X identity Y\end{tabular} & \begin{tabular}[c]{@{}l@{}}$s_x=s_y$\\ $e_x=e_y$\end{tabular}  \\\bottomline
\end{tabular}}
\end{table}
\footnotetext[4]{No TLink for Allen's overlap relation is present in TimeML, also concluded by \citet{UzZaman:2011}.}
\stepcounter{footnote}


As can be seen in the last column there are only two point-wise operations in the point-algebraic constraints: an order operation ($<$), and an equality operation ($=$). To model to what degree each point-wise constraint is met, we employ hinge losses, with a margin $m_{\tau}$, as shown in Eq. \ref{eq:pointwise_loss}.

\begin{figure}[h!]
\centering
\includegraphics[width=.47\textwidth]{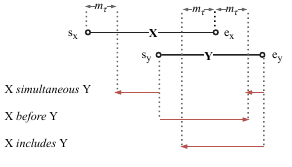}
\caption{\label{fig:loss_example} Visualization of the time-line loss $L_\tau$ with margin $m_{\tau}$, for two events X and Y, and TLinks \textit{simultaneous}, \textit{before}, and \textit{includes}. The red arrows' lengths indicate the loss per relation, i.e. how much the points should be shifted to satisfy each relation.}
\end{figure}

To explain the intuition and notation: If we have a point-wise expression $\xi$ of the form $x < y$ (first case of Eq. \ref{eq:pointwise_loss}), then the predicted point $\hat{x}$ should be at least a distance $m_{\tau}$ smaller (or earlier on the time-line) than predicted point $\hat{y}$ in order for the loss to be 0. Otherwise, the loss represents the distance $\hat{x}$ or $\hat{y}$ still has to move to make $\hat{x}$ smaller than $\hat{y}$ (and satisfy the constraint). For the second case, if $\xi$ is of the form $x = y$, then point $\hat{x}$ and $\hat{y}$ should lie very close to each other, i.e. at most a distance $m_{\tau}$ away from each other. Any distance further than the margin $m_{\tau}$ is counted as loss. Notice that if we set margin $m_{\tau}$ to 0, the second case becomes an L1 loss $|\hat{x} - \hat{y}|$. 
However, we use a small margin $m_{\tau}$ to promote some distance between ordered points and prevent confusion with equality.
Fig. \ref{fig:loss_example} visualizes the loss for three TLinks. 

\begin{align}
\label{eq:pointwise_loss}
L_p(\xi|t,\theta)  = 
\begin{cases}
\max(\hat{x} + m_{\tau} - \hat{y}, 0) &\text{ iff } x < y \\
\max(| \hat{x} - \hat{y} | - m_{\tau}, 0) &\text{ iff } x=y\\
\end{cases}
\end{align}

The total time-line loss $L_\tau(t|\theta)$ of a model with parameters $\theta$ on text $t$ with ground-truth TLinks $R(t)$, is the sum of the TLink-level losses of all TLinks $r \in R(t)$. Each TLink-level loss $L_r(r|t,\theta)$ for TLink $r$ is the sum of the point-wise losses $L_p(\xi|t,\theta)$ of the corresponding point-algebraic constraints $\xi \in I_{PA}(r)$ from Table \ref{tab:relations}.\footnote{The TLink \textit{during} and its inverse are mapped to \textit{simultaneous}, following the evaluation of TempEval-3.} 

\begin{align}
\label{eq:relation_loss}
L_r(r|t,\theta) &= \sum_{\xi \in I_{PA}(r)} L_p(\xi|t, \theta)\\
\label{eq:timeline_loss}
L_\tau(t, \theta) &= \sum_{r \in R(t)}  L_r(r|t,\theta)
\end{align}

\subsubsection*{Discriminative Time-line Losses}
To promote a more explicit difference between the relations on the time-line we introduce two discriminative loss functions, $L_{\tau ce}$ and $L_{\tau h}$, which build on top of $L_r$.
Both discriminative loss functions use an intermediate score $S(r|t, \theta)$ for each TLink $r$ based on the predicted time-line. As scoring function, we use the negative $L_r$ loss, as shown in Eq. \ref{eq:score}.
\begin{align}
\label{eq:score}
S(r|t,\theta) = -L_r(r|t,\theta)
\end{align}
Then, a lower time-line loss $L_r(r|t,\theta)$ results in a higher score for relation type $r$. Notice that the maximum score is 0, as this is the minimum $L_r$.

\subsubsection*{Probabilistic Loss ($L_{\tau ce}$)}
Our first discriminative loss is a cross-entropy based loss. For this the predicted scores are normalized using a soft-max over the possible relation types ($TL$). The resulting probabilities are used to calculate a cross-entropy loss, shown in Eq. \ref{eq:Ldce}. This way, the loss does not just promote the correct relation type but also distantiates from the other relation types.
\begin{align}
\label{eq:Ldce}
L_{\tau ce}(t| \theta) = - \sum_{r \in R(t)} r \cdot \log \Big( \frac{e^{S(r|t,\theta)}}{\sum_{r' \in TL} e^{S(r'|t,\theta)}}\Big)
\end{align}

\subsubsection*{Ranking Loss ($L_{\tau h}$)}
When interested in discriminating relations on the time-line, we want the correct relation type to have the highest score from all possible relation types $TL$. To represent this perspective, we also define a ranking loss with a score margin $m_h$ in Eq. \ref{eq:Ldh}.
\begin{multline}
\label{eq:Ldh}
L_{\tau h}(t| \theta) =\\ \sum_{r \in R(t)} \sum_{r' \in TL\setminus\{r\}} \max(S(r'|t,\theta) - S(r|t,\theta) + m_h, 0)
\end{multline}
\subsubsection*{Training Procedure}

S-TLM and C-TLM are trained by by iterating through the training texts, sampling mini-batches of 32 annotated TLinks. For each batch we (1) perform a forward pass, (2) calculate the total loss (for one of the loss functions), (3) derive gradients using Adam\footnote{Using the default parameters from the paper.} \cite{Kingma2014Adam:Optimization}, and (4) update the model parameters $\theta$ via back-propagation. After each epoch we shuffle the training texts. As stopping criteria we use early stopping \cite{morgan1990generalization}, with a patience of 100 epochs and a maximum number of 1000 epochs.

\subsection{From TLinks to Time-lines  (TL2RTL)}

To model the indirect route, we construct a novel method, TL2RTL, that predicts relative time lines from a subset of TLinks, shown in Fig \ref{fig:example}. One can choose any method to obtain a set of TLinks $R(t)$ from a text $t$, serving as input to TL2RTL. 
TL2RTL constructs a relative time-line, by assigning start and end values to each temporal entity, such that the resulting time-line satisfies the extracted TLinks $R(t)$ by minimizing a loss function that is 0 when the extracted TLinks are satisfied. TL2RTL on itself is a method and not a model. The only variables over which it optimizes the loss are the to be assigned starts and duration values.

In detail, for a text $t$, with annotated entities $E(t)$, we first extract a set of TLinks $R(t)$. In this work, to extract TLinks, we use the current state-of-the-art structured TLink extraction model by \citet{Ning2017}. Secondly, we assign a start variable $s_i$, and duration variable $d_i$ to each entity $i\in E(t)$. Similar to S-TLM and C-TLM, for each $i \in E(t)$, $d_i$ is bounded by a minimum duration $d_{min}$ to ensure start $s_i$ always lies before end $e_i$. Also, we model the DCT start $s_{\textsc{dct}}$ as a constant, and its duration $d_{\textsc{dct}}$ as a variable. Then we minimize one of the loss functions $L_\tau$, $L_{\tau ce}$, or $L_{\tau h}$ on the extracted TLinks $R(t)$, obtaining three TL2RTL variants, one for each loss. If the initially extracted set of TLinks $R(t)$ is consistent, and the loss is minimized sufficiently, all $s_i$ and $d_i$ form a relative time-line that satisfies the TLinks $R(t)$, but from which we can now also derive consistent TLinks for any entity pair, also the pairs that were not in $R(t)$.
To minimize the loss we use Adam for 10k epochs until the loss is zero for each document.\footnote{For some documents the extracted TLinks were temporally inconsistent, resulting in a non-zero loss. Nevertheless, $>96\%$ of the extracted TLinks were satisfied.}

\section{Experiments}

\subsection{Evaluation and Data}

Because prediction of relative time-lines trained on TimeML-style annotations is new, we cannot compare our model directly to relation extraction or classification models, as the latter do not provide completely temporally consistent TLinks for all possible entity pairs, like the relative time-lines do. Neither can we compare directly to existing absolute time-line prediction models such as \citet{Reimers2017EventClassifiers} because they are trained on different data with a very different annotation scheme.

To  evaluate the quality of the relative time-line models in a fair way, we use TimeML-style test sets as follows: (1) We predict a time-line for each test-text, and (2) we check for all ground-truth annotated TLinks that are present in the data, what would be the derived relation type based on the predicted time-line, which is the relation type that gives the lowest time-line loss $L_r$. This results in a TLink assignment for each annotated pair in the TimeML-style reference data, and therefor we can use similar metrics. As evaluation metric we employ the temporal awareness metric, used in TempEval-3, which takes into account temporal closure \cite{UzZaman2013}. Notice that although we use the same metric, comparison against relation classification systems would be unfair, as our model assigns consistent labels to all pairs, whereas relation classification systems do not.

For training and evaluation we use two data splits, TE$^\ddagger$ and TD$^\ddagger$, exactly following \citet{Ning2017}. Some statistics about the data are shown in Table \ref{tab:splits}.\footnote{We explicitly excluded all test documents from training as some corpora annotated the same documents.} The splits are constituted from various smaller datasets: the TimeBank (TB) \cite{Pustejovsky2002TheCorpus}, the AQUANT dataset (AQ), and the platinum dataset (PT) all from TempEval-3 \cite{UzZaman2013}. And, the TimeBank Dense \cite{Cassidy2014}
, and the Verb-Clause dataset (VC) \cite{Bethard2007}. 

\begin{table*}
\centering
\caption{\label{tab:splits} Dataset splits used for evaluation (indicated with $\ddagger$).}
\begin{tabular}{llcclcc}
\topline
\textbf{Split} & \textbf{Training data} & \textbf{\#TLinks} & \textbf{\#Documents} & \textbf{Test data} & \textbf{\#TLinks} & \textbf{\#Documents}\\\midline
TD$^\ddagger$ & TD (train+dev) & 4.4k & 27 & TD (test) & 1.3k & 9 \\
TE3$^\ddagger$ & TB, AQ, VC, TD (full) & 17.5k & 256 & PT & 0.9k & 20 \\
\bottomline
\end{tabular}
\end{table*}
\subsection{Hyper-parameters and Preprocessing}

\begin{table}[ht!]
\centering
\caption{\label{tab:hyperparameters}Hyper-parameters from the \mbox{experiments}.}
\begin{tabular}{lc}
\topline
\textbf{Hyper-parameter} & \textbf{Value} \\
\midline
Document-creation starting time ($s_{\textsc{dct}}$) & 0 \\
Minimum event duration ($d_{min}$) &0.1\\
Time-line margin ($m_{\tau}$) & 0.025\\
Hinge loss margin ($m_{h}$) & 0.1\\	
\midline
Dropout ($\alpha_d$) & 0.1\\
Word-level RNN units ($\alpha_{rnn}$) &25\\
Word-embedding size ($\alpha_{wemb}$) &50\\
POS-embedding size&10\\
\bottomline
\end{tabular} 
\end{table}

Hyper-parameters shared in all settings can be found in Table \ref{tab:hyperparameters}. The following hyper-parameters are tuned using grid search on a development set (union of TB and AQ): $d_{min}$ is chosen from $\{1, 0.1, 0.01\}$, $m_\tau$ from $\{0, 0.025, 0.05, 0.1\}$, $\alpha_d$ from $\{0, 0.1, 0.2, 0.4, 0.8\}$, and $\alpha_{rnn}$ from $\{10, 25, 50\}$. We use LSTM \cite{hochreiter1997long} as RNN units\footnote{We also experimented with GRU as RNN type, obtaining similar results.} and employ 50-dimensional GloVe word-embeddings pre-trained\footnote{https://nlp.stanford.edu/projects/glove} on 6B words (Wikipedia and NewsCrawl) to initialize the models' word embeddings. 

We use very simple tokenization and consider
punctuation\footnote{\texttt{$,./\backslash$"'=+-;:()!?<>\%\&\$*|[]\{\}}} or newline tokens as individual
tokens, and split on spaces. Additionally, we lowercase the text and use the Stanford POS Tagger \cite{toutanova2003feature} to obtain POS.

\section{Results}
We compared our three proposed models for the three loss functions $L_\tau$, $L_{\tau ce}$, and $L_{\tau h}$, and their linear (unweighted) combination $L_*$, on TE3$^\ddagger$ and TD$^\ddagger$, for which the results are shown in Table \ref{tab:results}.
\begin{table}[]
\centering
\caption{Evaluation of relative time-lines for each model and loss function, where $L_*$ indicates the (unweighted) sum of $L_\tau$, $L_{\tau ce}$, and $L_{\tau h}$.}
\label{tab:results}
\resizebox{.49\textwidth}{!}{
\begin{tabular}{llcccccc}
\topline 
 & \multicolumn{3}{c}{\bf TE3$^\ddagger$ }  & \multicolumn{3}{c}{\bf TD$^\ddagger$ }\\
\textbf{Model}     &\textbf{P}&\textbf{R}&\textbf{F} &\textbf{P}&\textbf{R}&\textbf{F}\\\midline


\textit{Indirect: $O(n^2)$}\\
\hspace{.3cm}TL2RTL ($L_{\tau }$)                   & 53.5 & 51.1 & 52.3 & 59.1 & 61.2 & 60.1\\
\hspace{.3cm}TL2RTL ($L_{\tau ce}$)                   & 53.9 & 51.7 & \textbf{52.8} & 61.2 & 60.7 & 60.9\\
\hspace{.3cm}TL2RTL ($L_{\tau h}$)                   & 52.8 & 51.1 & 51.9 & 57.9 & 60.6 & 59.2\\
\hspace{.3cm}TL2RTL ($L_{*}$)                   & 52.6 & 52.0 & 52.3 & 62.3 & 62.3 & \textbf{62.3}\\
\midline
\textit{Direct: $O(n)$}\\

\hspace{.3cm}S-TLM ($L_{\tau }$)                  & 50.1 & 50.4 & 50.2 & 57.8 & 59.5 & \textbf{58.6}\\
\hspace{.3cm}S-TLM ($L_{\tau ce}$)                  & 50.1 & 50.0 & 50.1 & 53.4 & 53.5 & 53.5\\
\hspace{.3cm}S-TLM ($L_{\tau h}$)                  & 51.5 & 51.7 & \textbf{51.6} & 55.1 & 56.4 & 55.7\\
\hspace{.3cm}S-TLM ($L_{*}$)                   & 50.9 & 51.0 & 51.0 & 56.5 & 55.3 & 55.9
\vspace{.1cm}\\
\hspace{.3cm}C-TLM ($L_{\tau }$)                  & 56.2 & 56.1 & \textbf{56.1} & 57.1 & 59.7 & \textbf{58.4}\\
\hspace{.3cm}C-TLM ($L_{\tau ce}$)                 & 54.4 & 55.4 & 54.9 & 52.4 & 57.3 & 54.7\\
\hspace{.3cm}C-TLM ($L_{\tau h}$)                  & 55.7 & 55.5 & 55.6 & 55.3 & 54.9 & 55.1\\
\hspace{.3cm}C-TLM ($L_{*}$)                   & 54.0 & 54.3 & 54.1 & 54.6 & 53.5 & 54.1\\
\bottomline
\end{tabular}}
\end{table}

A trend that can be observed is that overall performance on TD$^\ddagger$ is higher than that of TE3$^\ddagger$, even though less documents are used for training. We inspected why this is the case, and this is caused by a difference in class balance between both test sets. In TE3$^\ddagger$ there are many more TLinks of type \textit{simultaneous} (12\% versus 3\%), which are very difficult to predict, resulting in lower scores for TE3$^\ddagger$ compared to TD$^\ddagger$. The difference in performance between the datasets is probably also be related to the dense annotation scheme of TD$^\ddagger$ compared to the sparser annotations of TE3$^\ddagger$, as dense annotations give a more complete temporal view of the training texts. 
For TL2RTL better TLink extraction\footnote{F1 of 40.3 for TE3$^\ddagger$ and 48.5 for TD$^\ddagger$ \cite{Ning2017}} is also propagated into the final time-line quality. 

If we compare loss functions $L_\tau$, $L_{\tau ce}$, and $L_{\tau h}$, and combination $L_*$, it can be noticed that, although all loss functions seem to give fairly similar performance, $L_\tau$ gives the most robust results (never lowest), especially noticeable for the smaller dataset TD$^\ddagger$. This is convenient, because $L_\tau$ is fastest to compute during training, as it requires no score calculation for each TLink type. $L_\tau$ is also directly interpretable on the time-line. The combination of losses $L_*$ shows mixed results, and has lower performance for S-TLM and C-TLM, but better performance for TL2RTL. However, it is slowest to compute, and less interpretable, as it is a combined loss.


Moreover, we can clearly see that on TE3$^\ddagger$, C-TLM performs better than the indirect models, across all loss functions. This is a very interesting result, as C-TLM is an order of complexity faster in prediction speed compared to the indirect models ($O(n)$ compared to $O(n^2)$ for a text with $n$ entities).\footnote{We do not directly compare prediction speed, as it would result in unfair evaluation because of implementation differences. However, currently, C-TLM predicts at $\sim$100 w/s incl. POS tagging, and $\sim$2000 w/s without. When not using POS, overall performance decreases consistently with 2-4 points.} We further explore why this is the case through our error analysis in the next section. 

\begin{figure*}[th!]
\centering
\begin{tabular}{c c c c c c }
\topline
 & \textbf{B}& \textbf{A} & \textbf{II} & \textbf{I} & \textbf{S}  \\\topline
\textbf{B} & 24.8\% \cellcolor[gray]{.8}& 4.7\% & 2.8\% & 1.6\% & 0.1\% \\
\textbf{A} & 5.0\% & 15.8\% \cellcolor[gray]{.8} & 3.2\% & 0.5\% & 0.0\% \\
\textbf{II} & 3.2\% & 3.2\% & 13.0\% \cellcolor[gray]{.8} & 0.6\% & 0.1\% \\
\textbf{I} & 4.0\% & 1.2\% & 1.0\% & 3.2\% \cellcolor[gray]{.8} & 0.0\% \\
\textbf{S} & 4.4\% & 3.0\% & 2.6\% & 1.3\% & 0.4\% \cellcolor[gray]{.8} \\
\bottomline
\end{tabular}
\hfill
\begin{tabular}{c c c c c c }
\topline
 & \textbf{B}& \textbf{A} & \textbf{II} & \textbf{I} & \textbf{S}  \\\topline
\textbf{B} & 23.0\% \cellcolor[gray]{.8}& 8.2\% & 1.3\% & 0.9\% & 0.8\% \\
\textbf{A} & 4.7\% & 17.1\% \cellcolor[gray]{.8} & 1.8\% & 0.3\% & 0.5\% \\
\textbf{II} & 4.3\% & 4.4\% & 11.1\% \cellcolor[gray]{.8} & 0.4\% & 0.0\% \\
\textbf{I} & 1.6\% & 5.4\% & 0.5\% & 1.3\% \cellcolor[gray]{.8} & 0.5\% \\
\textbf{S} & 4.3\% & 4.1\% & 1.8\% & 0.6\% & 0.9\% \cellcolor[gray]{.8} \\
\bottomline
\end{tabular}
\caption{\label{fig:confusion} On the \textbf{left}, the confusion matrix of C-TLM ($L_{\tau }$), and on the \textbf{right} of TL2RTL ($L_{\tau ce}$), on TE3$^\ddagger$ for the top-5 most-frequent TLinks (together 95\% of data): \textsc{before} (B), \textsc{after} (A), \textsc{is included} (II), \textsc{includes} (I), and \textsc{simultaneous} (S). Predictions are shown on the x-axis and ground-truth on the y-axis.}
\end{figure*}

On TD$^\ddagger$, the indirect models seem to perform slightly better. We suspect that the reason for this is that C-TLM has more parameters (mostly the LSTM weights), and thus requires more data (TD$^\ddagger$ has much fewer documents than TE3$^\ddagger$) compared to the indirect methods. Another result supporting this hypothesis is the fact that the difference between C-TLM and S-TLM is small on the smaller TD$^\ddagger$, indicating that C-TLM does not yet utilize contextual information from this dataset, whereas, in contrast, on TE3$^\ddagger$, the larger dataset, C-TLM clearly outperforms S-TLM across all loss functions, showing that when enough data is available C-TLM learns good LSTM weights that exploit context substantially.








 \section{Error Analysis}
We compared predictions of TL2RTL($L_{\tau }$) with those of C-TLM ($L_{\tau}$), the best models of each paradigm. In Table \ref{fig:confusion}, we show the confusion matrices of both systems on TE3$^\ddagger$.

When looking at the overall pattern in errors, both models seem to make similar confusions on both datasets (TD$^\ddagger$ was excluded for space constraints). 
Overall, we find that \textit{simultaneous} is the most violated TLink for both models. This can be explained by two reasons: (1) It is the least frequent TLink in both datasets. And (2), simultaneous entities are often co-referring events. Event co-reference resolution is a very difficult task on its own.
 
We also looked at the average token-distance between arguments of correctly satisfied TLinks by the time-lines of each model. For TL2RTL ($L_{\tau }$) this is 13 tokens, and for C-TLM ($L_{\tau}$) 15. When looking only at the TLinks that C-TLM ($L_{\tau}$) satisfied and TL2RTL ($L_{\tau }$) did not, the average distance is 21. These two observations suggest that the direct C-TLM ($L_{\tau}$) model is better at positioning entities on the time-line that lie further away from each other in the text. An explanation for this can be error propagation of TLink extraction to the time-line construction, as the pairwise TLink extraction of the indirect paradigm extracts TLinks in a contextual window, to prune the $O(n^2)$ number of possible TLink candidates. This consequently prevents TL2RTL to properly position distant events with respect to each other.



To get more insight in what the model learns we calculated mean durations and mean starts of C-TLM ($L_\tau$) predictions. Table \ref{tab:sd_examples} contains examples from the top-shortest, and top-longest duration assignments and earliest and latest starting points. We observe that events that generally have more events included are assigned longer duration and vice versa. And, events with low start values are in the past tense and events with high start values are generally in the present (or future) tense.

\begin{table}
\centering
\caption{\label{tab:sd_examples} Example events from the top-shortest/longest durations and top-earliest/latest start values assigned by the model.}
\begin{tabular}{llll}
\topline
\textbf{Short $d$} & \textbf{Long $d$} & \textbf{Early $s$} & \textbf{Late $s$}\\
\midline
started & going  &destroyed &realize \\
meet & expects &finished &bring \\
 entered & recession &invaded &able \\
told & war &pronounced &got \\
arrived & support &created &work \\
allow & make &took &change \\
send & think &appeared &start \\
asked & created &leaving &reenergize \\
\bottomline
\end{tabular}
\end{table}



\section{Discussion}
A characteristic of our model is that it assumes that all events can be placed on a single time-line, and that it does not assume that unlabeled pairs are temporally unrelated. This has big advantages: it results in fast prediction, and missed annotation do not act as noise to the training, as they do for pairwise models. \citet{P18-1122} argue that actual, negated, hypothesized, expected or opinionated events should possibly be annotated on separate time-axis. We believe such multi-axis representations can be inferred from the generated single time-lines if hedging information is recognized.


\section{Conclusions}

This work leads to the following three main contributions\footnote{Code is available at: \href{http://liir.cs.kuleuven.be/software_pages/relative_timelines.php}{liir.cs.kuleuven.be/software.php}}: 
(1) Three new loss functions that connect the interval-based TimeML-annotations to points on a time-line, (2) A new method, TL2RTL, to predict relative time-lines from a set of predicted temporal relations. And (3), most importantly, two new models, S-TLM and C-TLM, that -- to our knowledge for the first time -- predict (relative) time-lines in linear complexity from text, by evading the computationally expensive (often $O(n^2)$) intermediate relation extraction phase in earlier work. 
From our experiments, we conclude that the proposed loss functions can be used effectively to train direct and indirect relative time-line models, and that, when provided enough data, the -- much faster -- direct model C-TLM outperforms the indirect method TL2RTL.

As a direction for future work, it would be very interesting to extend the current models, diving further into direct time-line models, and learn to predict absolute time-lines, i.e. making the time-lines directly mappable to calender dates and times, e.g. by exploiting complementary data sources such as the EventTimes Corpus \cite{Reimers2016a} and extending the current loss functions accordingly. The proposed models also provide a good starting point for research into probabilistic time-line models, that additionally model the (un)certainty of the predicted positions and durations of the entities.

\section*{Acknowledgments}
The authors thank Geert Heyman and the reviewers for their constructive comments which helped us to improve the paper. This work was funded by the KU Leuven C22/15/16 project "MAchine Reading of patient recordS (MARS)", and by the IWT-SBO 150056 project "ACquiring CrUcial Medical information Using LAnguage TEchnology" (ACCUMULATE).


\bibliography{main}

\begin{thebibliography}{32}
\expandafter\ifx\csname natexlab\endcsname\relax\def\natexlab#1{#1}\fi

\bibitem[{Allen(1990)}]{Allen}
James~F Allen. 1990.
\newblock Maintaining knowledge about temporal intervals.
\newblock \emph{Readings in Qualitative Reasoning about Physical Systems},
  pages 361--372.

\bibitem[{Bethard(2013)}]{bethard2013cleartk}
Steven Bethard. 2013.
\newblock {ClearTK}-{TimeML}: A minimalist approach to tempeval 2013.
\newblock In \emph{Proceedings of SemEval}, volume~2, pages 10--14. ACL.

\bibitem[{Bethard et~al.(2007)Bethard, Martin, and Klingenstein}]{Bethard2007}
Steven Bethard, James~H. Martin, and Sara Klingenstein. 2007.
\newblock {Timelines from text: Identification of syntactic temporal
  relations}.
\newblock In \emph{Proceedings of ICSC}, pages 11--18.

\bibitem[{Bethard et~al.(2016)Bethard, Savova, Chen, Derczynski, Pustejovsky,
  and Verhagen}]{bethard2016semeval}
Steven Bethard, Guergana Savova, Wei-Te Chen, Leon Derczynski, James
  Pustejovsky, and Marc Verhagen. 2016.
\newblock Semeval-2016 task 12: Clinical tempeval.
\newblock In \emph{Proceedings of SemEval}, pages 1052--1062. ACL.

\bibitem[{Bethard et~al.(2017)Bethard, Savova, Palmer, and
  Pustejovsky}]{Bethard2017SemEval-2017TempEval}
Steven Bethard, Guergana Savova, Martha Palmer, and James Pustejovsky. 2017.
\newblock {SemEval-2017 Task 12: Clinical TempEval}.
\newblock In \emph{Proceedings of SemEval}, pages 565--572. ACL.

\bibitem[{Cassidy et~al.(2014)Cassidy, McDowell, Chambers, and
  Bethard}]{Cassidy2014}
Taylor Cassidy, Bill McDowell, Nathanael Chambers, and Steven Bethard. 2014.
\newblock An annotation framework for dense event ordering.
\newblock In \emph{Proceedings of ACL}, pages 501--506. ACL.

\bibitem[{Chambers et~al.(2014)Chambers, Cassidy, McDowell, and
  Bethard}]{chambers2014dense}
Nathanael Chambers, Taylor Cassidy, Bill McDowell, and Steven Bethard. 2014.
\newblock Dense event ordering with a multi-pass architecture.
\newblock \emph{Transactions of the Association for Computational Linguistics},
  2:273--284.

\bibitem[{Chambers and Jurafsky(2008)}]{chambers2008jointly}
Nathanael Chambers and Dan Jurafsky. 2008.
\newblock Jointly combining implicit constraints improves temporal ordering.
\newblock In \emph{Proceedings of EMNLP}, pages 698--706. ACL.

\bibitem[{Cheng and Miyao(2017)}]{cheng2017classifying}
Fei Cheng and Yusuke Miyao. 2017.
\newblock Classifying temporal relations by bidirectional {LSTM} over
  dependency paths.
\newblock In \emph{Proceedings of ACL}, volume~2, pages 1--6. ACL.

\bibitem[{Denis and Muller(2011)}]{Denis2011}
Pascal Denis and Philippe Muller. 2011.
\newblock {Predicting globally-coherent temporal structures from texts via
  endpoint inference and graph decomposition}.
\newblock In \emph{Proceedings of IJCAI}, pages 1788--1793.

\bibitem[{Derczynski(2017)}]{Derczynski2017}
Leon~RA Derczynski. 2017.
\newblock \emph{Automatically Ordering Events and Times in Text}, volume 677.
\newblock Springer.

\bibitem[{Dligach et~al.(2017)Dligach, Miller, Lin, Bethard, and
  Savova}]{dligach2017neural}
Dmitriy Dligach, Timothy Miller, Chen Lin, Steven Bethard, and Guergana Savova.
  2017.
\newblock Neural temporal relation extraction.
\newblock In \emph{Proceedings of EACL}, volume~2, pages 746--751.

\bibitem[{Do et~al.(2012)Do, Lu, and Roth}]{Do2012JointConstruction}
Quang~Xuan Do, Wei Lu, and Dan Roth. 2012.
\newblock Joint inference for event timeline construction.
\newblock In \emph{Proceedings of EMNLP-CoNLL}, pages 677--687. ACL.

\bibitem[{Hochreiter and Schmidhuber(1997)}]{hochreiter1997long}
Sepp Hochreiter and J{\"u}rgen Schmidhuber. 1997.
\newblock Long short-term memory.
\newblock \emph{Neural Computation}, 9(8):1735--1780.

\bibitem[{Kingma and Ba(2014)}]{Kingma2014Adam:Optimization}
Diederik~P Kingma and Jimmy Ba. 2014.
\newblock Adam: A method for stochastic optimization.
\newblock \emph{arXiv preprint arXiv:1412.6980}.

\bibitem[{Laokulrat et~al.(2015)Laokulrat, Miwa, and Tsuruoka}]{Laokulrat2015}
Natsuda Laokulrat, Makoto Miwa, and Yoshimasa Tsuruoka. 2015.
\newblock Stacking approach to temporal relation classification.
\newblock \emph{Journal of Natural Language Processing}, 22(3):171--196.

\bibitem[{Leeuwenberg and Moens(2017)}]{Leeuwenberg2017a}
Artuur Leeuwenberg and Marie-Francine Moens. 2017.
\newblock {Structured learning for temporal relation extraction from clinical
  records}.
\newblock In \emph{Proceedings of EACL}, volume~1, pages 1150--1158. ACL.

\bibitem[{Lin et~al.(2015)Lin, Dligach, Miller, Bethard, and
  Savova}]{lin2015multilayered}
Chen Lin, Dmitriy Dligach, Timothy~A Miller, Steven Bethard, and Guergana~K
  Savova. 2015.
\newblock Multilayered temporal modeling for the clinical domain.
\newblock \emph{Journal of the American Medical Informatics Association},
  23(2):387--395.

\bibitem[{Mani et~al.(2006)Mani, Verhagen, Wellner, Lee, and
  Pustejovsky}]{Mani2006MachineRelations}
Inderjeet Mani, Marc Verhagen, Ben Wellner, Chong~Min Lee, and James
  Pustejovsky. 2006.
\newblock Machine learning of temporal relations.
\newblock In \emph{Proceedings of COLING-ACL}, pages 753--760. ACL.

\bibitem[{Mirza and Tonelli(2016)}]{Mirza2016a}
Paramita Mirza and Sara Tonelli. 2016.
\newblock {CATENA} : Causal and temporal relation extraction from natural
  language texts.
\newblock \emph{Proceedings of COLING}, pages 64--75.

\bibitem[{Morgan and Bourlard(1990)}]{morgan1990generalization}
Nelson Morgan and Herv{\'e} Bourlard. 1990.
\newblock Generalization and parameter estimation in feedforward nets: Some
  experiments.
\newblock In \emph{Advances in Neural Information Processing Systems}.

\bibitem[{Ning et~al.(2017)Ning, Feng, and Roth}]{Ning2017}
Qiang Ning, Zhili Feng, and Dan Roth. 2017.
\newblock A structured learning approach to temporal relation extraction.
\newblock \emph{Proceedings of EMNLP}, pages 1038--1048.

\bibitem[{Ning et~al.(2018)Ning, Wu, and Roth}]{P18-1122}
Qiang Ning, Hao Wu, and Dan Roth. 2018.
\newblock A multi-axis annotation scheme for event temporal relations.
\newblock In \emph{Proceedings of ACL}, pages 1318--1328. ACL.

\bibitem[{Pustejovsky et~al.(2002)Pustejovsky, Hanks, Saur{\'{i}}, See,
  Gaizauskas, Setzer, Sundheim, Day, Ferro, and
  {Dragomir}}]{Pustejovsky2002TheCorpus}
James Pustejovsky, Patrick Hanks, Roser Saur{\'{i}}, Andrew See, Robert
  Gaizauskas, Andrea Setzer, Beth Sundheim, David Day, Lisa Ferro, and
  {Dragomir}. 2002.
\newblock {The TIMEBANK Corpus}.
\newblock \emph{Natural Language Processing and Information Systems},
  4592:647--656.

\bibitem[{Reimers et~al.(2016)Reimers, Dehghani, and Gurevych}]{Reimers2016a}
Nils Reimers, Nazanin Dehghani, and Iryna Gurevych. 2016.
\newblock Temporal anchoring of events for the timebank corpus.
\newblock \emph{Proceedings of ACL}, pages 2195--2204.

\bibitem[{Reimers et~al.(2018)Reimers, Dehghani, and
  Gurevych}]{Reimers2017EventClassifiers}
Nils Reimers, Nazanin Dehghani, and Iryna Gurevych. 2018.
\newblock Event time extraction with a decision tree of neural classifiers.
\newblock \emph{Transactions of the Association for Computational Linguistics},
  6:77--89.

\bibitem[{Tourille et~al.(2017)Tourille, Ferret, Neveol, and
  Tannier}]{Tourille2017}
Julien Tourille, Olivier Ferret, Aurelie Neveol, and Xavier Tannier. 2017.
\newblock Neural architecture for temporal relation extraction: A {Bi-LSTM}
  approach for detecting narrative containers.
\newblock In \emph{Proceedings of ACL}, pages 224--230.

\bibitem[{Toutanova et~al.(2003)Toutanova, Klein, Manning, and
  Singer}]{toutanova2003feature}
Kristina Toutanova, Dan Klein, Christopher~D Manning, and Yoram Singer. 2003.
\newblock Feature-rich part-of-speech tagging with a cyclic dependency network.
\newblock In \emph{Proceedings of NAACL-HLT}, pages 173--180. ACL.

\bibitem[{UzZaman and Allen(2011)}]{UzZaman:2011}
Naushad UzZaman and James~F. Allen. 2011.
\newblock Temporal evaluation.
\newblock In \emph{Proceedings of ACL}, HLT '11, pages 351--356, Stroudsburg,
  PA, USA. ACL.

\bibitem[{UzZaman et~al.(2013)UzZaman, Llorens, Derczynski, Verhagen, Allen,
  and Pustejovsky}]{UzZaman2013}
Naushad UzZaman, Hector Llorens, Leon Derczynski, Marc Verhagen, James Allen,
  and James Pustejovsky. 2013.
\newblock {Semeval-2013 task 1: Tempeval-3: Evaluating time expressions,
  events, and temporal relations}.
\newblock \emph{Second joint conference on lexical and computational semantics
  (* SEM)}, 2:1--9.

\bibitem[{Vilain et~al.(1990)Vilain, Kautz, and
  Van~Beek}]{vilain1990constraint}
Marc Vilain, Henry Kautz, and Peter Van~Beek. 1990.
\newblock Constraint propagation algorithms for temporal reasoning: A revised
  report.
\newblock In \emph{Readings in Qualitative Reasoning about Physical Systems},
  pages 373--381. Elsevier.

\bibitem[{Yoshikawa et~al.(2009)Yoshikawa, Riedel, Asahara, and
  Matsumoto}]{yoshikawa2009jointly}
Katsumasa Yoshikawa, Sebastian Riedel, Masayuki Asahara, and Yuji Matsumoto.
  2009.
\newblock Jointly identifying temporal relations with markov logic.
\newblock In \emph{Proceedings of ACL-IJCNLP}, pages 405--413. ACL.

\end{thebibliography}
\bibliographystyle{acl_natbib_nourl}
\end{document}